\documentclass[11pt]{article}

\usepackage{acl}
\usepackage{times}
\usepackage{latexsym}
\usepackage{amsmath}
\usepackage{amssymb}
\usepackage{multirow}
\usepackage{graphicx}
\usepackage{algorithm}
\usepackage{algpseudocode}

\usepackage[T1]{fontenc}

\usepackage[utf8]{inputenc}

\usepackage{microtype}

\usepackage{inconsolata}

\usepackage{graphicx}

\title{FocalOrder: Focal Preference Optimization for Reading Order Detection}

\author{
    Fuyuan Liu\textsuperscript{1}\thanks{\ \ Equal contribution.} ,
    Dianyu Yu\textsuperscript{1,3}\footnotemark[1] ,
    He Ren\textsuperscript{1},
    Nayu Liu\textsuperscript{4},
    Xiaomian Kang\textsuperscript{2},
    Delai Qiu\textsuperscript{1},
    Fa Zhang\textsuperscript{1}, \\
    \textbf{Genpeng Zhen}\textsuperscript{1},
    \textbf{Shengping Liu}\textsuperscript{1},
    \textbf{Jiaen Liang}\textsuperscript{1},
    \textbf{Wei Huang}\textsuperscript{1},
    \textbf{Yining Wang}\textsuperscript{1}\thanks{\ \ Corresponding author.},
    \textbf{Junnan Zhu}\textsuperscript{2}\footnotemark[2] \\
    \\
    \textsuperscript{1}Unisound AI Technology Co.Ltd \\
    \textsuperscript{2}MAIS, Institute of Automation, Chinese Academy of Sciences \\
    \textsuperscript{3}Beihang University \\
    \textsuperscript{4}School of Computer Science and Technology, Tiangong University \\
    \\
    \texttt{junnan.zhu@nlpr.ia.ac.cn, wangyining@unisound.com}
}

\begin{document}
\maketitle

\begin{abstract}
Reading order detection is the foundation of document understanding.
Most existing methods rely on uniform supervision, implicitly assuming a constant difficulty distribution across layout regions. In this work, we challenge this assumption by revealing a critical flaw: \textbf{Positional Disparity}, a phenomenon where models demonstrate mastery over the deterministic start and end regions but suffer a performance collapse in the complex intermediate sections.
This degradation arises because standard training allows the massive volume of easy patterns to drown out the learning signals from difficult layouts.
To address this, we propose \textbf{FocalOrder}, a framework driven by \textbf{Focal Preference Optimization (FPO)}.
Specifically, FocalOrder employs adaptive difficulty discovery with exponential moving average mechanism to dynamically pinpoint hard-to-learn transitions, while introducing a difficulty-calibrated pairwise ranking objective to enforce global logical consistency.
Extensive experiments demonstrate that FocalOrder establishes new state-of-the-art results on OmniDocBench v1.0 and Comp-HRDoc.
Our compact model not only outperforms competitive specialized baselines but also significantly surpasses large-scale general VLMs.
These results demonstrate that aligning the optimization with intrinsic structural ambiguity of documents is critical for mastering complex document structures.
\end{abstract}

\section{Introduction}
\label{sec:intro}

\begin{figure}[t]
  \centering
  \setlength{\abovecaptionskip}{0.12cm}
  \setlength{\belowcaptionskip}{-0.5cm}
  \includegraphics[width=\columnwidth]{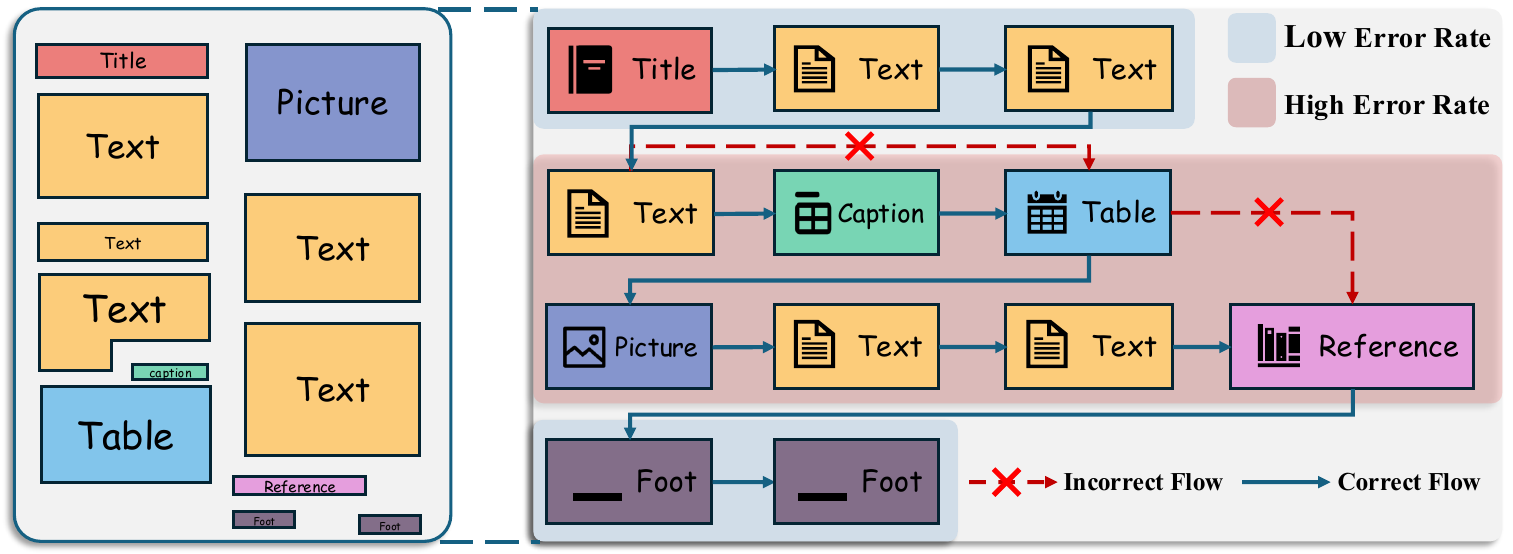}
  \caption{Illustration of Positional Disparity. While representative models demonstrate mastery over deterministic regions (start/end), they suffer from significant performance degradation in the document body. This reveals a misalignment between the uniform supervision used in training and the non-uniform structural complexity of real-world documents.}
  \label{fig:picture1}
\end{figure}

Recently, document intelligence has evolved from simple optical character recognition to complex semantic and structural understanding~\cite{cui2021document,ke2025large}. Reading order detection serializes spatially scattered regions into a coherent logical flow~\cite{giovannini2025survey}. It serves as the cognitive backbone for downstream applications, ranging from Retrieval-Augmented Generation (RAG)~\cite{zhang2025ocr} to complex logical reasoning~\cite{MathewKJ21}.

Recent advancements have transitioned from traditional discriminative models~\cite{meunier2005optimized} to end-to-end generative pipelines~\cite{wang2021layoutreader,mineru2.5}.
However, a fundamental gap remains between \textit{how documents are structured} and \textit{how models are optimized}.

Standard approaches predominantly rely on \textbf{uniform supervision}, such as standard Cross-Entropy.
This method implicitly assumes that the difficulty of predicting the next layout element is constant throughout the document.
By conducting a rigorous empirical analysis across diverse architectures (Section~\ref{sec:analysis}), we uncover a systematic bias called \textbf{Positional Disparity}.
As illustrated in Figure~\ref{fig:picture1}, models achieve near-perfect accuracy in low-entropy regions like titles and references that follow rigid templates.
In contrast, they suffer a catastrophic performance drop in the intermediate sections of the document body.
This suggests that current optimization objectives are dominated by the massive volume of trivial and deterministic transitions, which drowns out the learning signals for complex regions.
As a result, the model effectively ``memorizes'' the templates at the boundaries while failing to learn the robust spatial reasoning required for the ambiguous layouts in the middle.

To bridge this optimization gap, we propose \textbf{FocalOrder}, a framework that shifts from uniform sequence modeling to an adaptive, curriculum-style optimization. To realize this strategy, we introduce \textbf{Focal Preference Optimization (FPO)}, a mechanism designed to dynamically align supervision intensity with layout ambiguity. Instead of treating all layout transitions equally, FocalOrder acknowledges that not all transitions are created equal.
Our approach consists of two complementary mechanisms designed to realize this focal strategy.
First, we introduce \textbf{Adaptive Difficulty Discovery}.
This mechanism uses an Exponential Moving Average (EMA) to track historical error rates.
It autonomously identifies structural bottlenecks where the model struggles, thereby determining \textit{where} the model needs to focus.
Second, we propose a \textbf{Difficulty-Calibrated Pairwise Ranking} objective.
Unlike standard contrastive losses, this module constructs preference pairs weighted by the discovered topological complexity.
It explicitly amplifies the learning signals from hard samples and forces the model to prioritize global logical coherence over local pattern matching.

We validate FocalOrder on comprehensive benchmarks, including the OmniDocBench (v1.0 and v1.5)~\cite{ouyang2024omnidocbench} and Comp-HRDoc~\cite{wang2024detect}.
Without introducing additional training data or scaling up parameters, FocalOrder establishes new state-of-the-art results on OmniDocBench v1.0 and Comp-HRDoc.
It effectively flattens the ``Inverted-U'' error curve.
Our findings demonstrate that the key to mastering complex layouts lies not just in larger architectures.
It lies in aligning the optimization landscape with the intrinsic entropy distribution of documents.

Our contributions are summarized as follows:
\begin{itemize}
    \item We identify and formalize \textit{Positional Disparity}. We reveal that standard uniform optimization fails to capture the varying complexity of document layouts.
    \item We propose FocalOrder, a novel framework incorporating Adaptive Difficulty Discovery and Difficulty-Calibrated Pairwise Ranking. This framework dynamically aligns the learning focus with structural ambiguity.
    \item Extensive experiments show that FocalOrder significantly reduces sorting errors in complex intermediate regions. It establishes new state-of-the-art performance on OmniDocBench v1.0 and Comp-HRDoc.
\end{itemize}

\section{Related Work}
\label{sec:related_work}

\textbf{Local Discriminative Models.}
Early research primarily treats reading order detection as a local classification problem.
These methods focused on predicting the relationship between pairs of text segments.
For instance, \citet{wu2008machine} employ SVMs to determine if one segment should precede another.
Later, graph neural networks (GNNs)~\cite{li2020end} are introduced to model the connectivity between neighboring regions.
While these approaches capture local geometric cues effectively, they often lack a global view of the document structure, requiring complex heuristics to assemble predictions.
To mitigate this limitation, recent work like MLARP~\cite{qiao2024reading} introduces global graph constraints to regularize binary relation predictions.
However, constructing a sequence from discrete relations remains a multi-stage process.

\textbf{Generative Sequence Models.}
To achieve global coherence, the field has shifted towards end-to-end sequence generation.
LayoutReader~\cite{wang2021layoutreader} pioneers this direction by formulating the task as a sequence-to-sequence problem, using attention mechanisms to predict the order of all regions globally.
Similarly, MonkeyOCR~\cite{monkeyocr} adopts this methodology for reading order.
Building on this, PaddleOCR-VL~\cite{cui2025paddleocr} incorporates pointer networks.
This architecture separates the sorting process from content recognition, improving stability.
More recently, systems like MinerU 2.5~\cite{mineru2.5} and dots.ocr~\cite{li2025dots} have adopted decoupled pipelines, explicitly predicting the reading order before text recognition to handle high-resolution documents better.
These generative methods have become the mainstream choice because they learn global dependencies directly from data.

\textbf{Limitations and The Optimization Gap.}
Despite the architectural advancements from local to global models, a fundamental limitation remains in how these models are optimized.
Almost all existing approaches, including the SOTA generative models, rely on uniform supervision (e.g., standard cross-entropy loss).
This training objective treats every step in the sequence as equally difficult.
It penalizes a mistake in a simple header just as heavily as a mistake in a complex nested table.
Although some recent studies like Infinity-Parser~\cite{wang2025infinity} and DeepSeek-OCR~\cite{deepseekocr} have attempted to use Reinforcement Learning (RL) to enforce structural constraints, they often suffer from training instability and sparse rewards.
In contrast to existing approaches, we argue that the core problem lies in the mismatch between uniform supervision and the uneven difficulty of document layouts.
Therefore, our work proposes a focal optimization framework.
Instead of treating all data equally, we dynamically identify and prioritize the ambiguous transitions in the document body, ensuring the model focuses on the most challenging parts of the structure.

\section{Analysis of Positional Disparity}
\label{sec:analysis}

Does the model predict equally well at all positions? To investigate the reliability of uniform supervision, we conduct a systematic empirical analysis on OmniDocBench and Comp-HRDoc. To ensure the universality of our findings, we evaluate multiple representative models, including LayoutReader~\cite{wang2021layoutreader}, PaddleOCR-VL~\cite{cui2025paddleocr}, and MinerU 2.5~\cite{mineru2.5}. We quantify the prediction error rate relative to the normalized document position. We map the sequence index $t$ of a document with length $T$ to a relative position $p = t/T \in [0, 1]$ and calculate the average error rate for each percentile bin.

\begin{figure}[t]
  \centering
  \setlength{\abovecaptionskip}{0.12cm}
  \setlength{\belowcaptionskip}{-0.1cm}
  \includegraphics[width=\columnwidth]{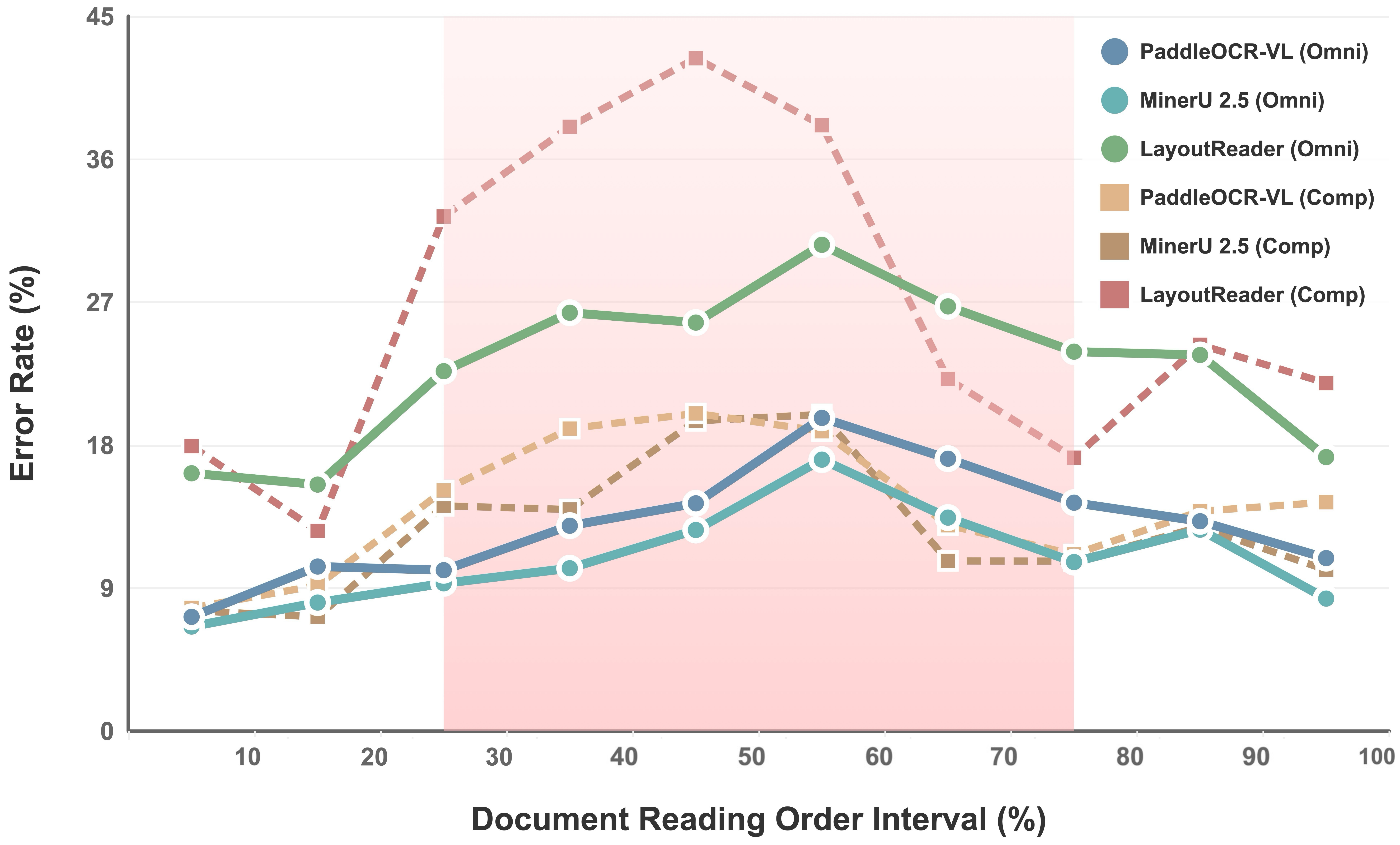}
  \caption{Error rates of representative models across normalized document positions. The consistent ``Inverted-U'' curve across datasets (solid lines: OmniDocBench, dashed lines: Comp-HRDoc) reveals a systematic bias, i.e., models struggle to serialize the complex document body compared to the rigid start and end templates.}
  \label{fig:Error}
\end{figure}

As shown in Figure~\ref{fig:Error}, all evaluated models exhibit a systematic bias termed Positional Disparity, characterized by a distinct ``Inverted-U'' error curve:
\begin{itemize}
    \item \textbf{Robust Start and End:} The initial and final segments of documents typically follow deterministic formatting templates, such as headers or references. Consequently, all models demonstrate robust mastery in these low-entropy regions.
    \item \textbf{Degradation in the Intermediate Sections:} In contrast, a pronounced increase in error rate is consistently observed within the document body (relative positions $20\%$--$80\%$). We hypothesize that this degradation stems from Structural Ambiguity, where the logical reading order deviates from simple geometric proximity. This pattern is most prevalent in the dense content of the document body.
\end{itemize}

\begin{figure}[t]
  \centering
  \setlength{\abovecaptionskip}{0.12cm}
  \setlength{\belowcaptionskip}{-0.3cm}
  \includegraphics[width=\columnwidth]{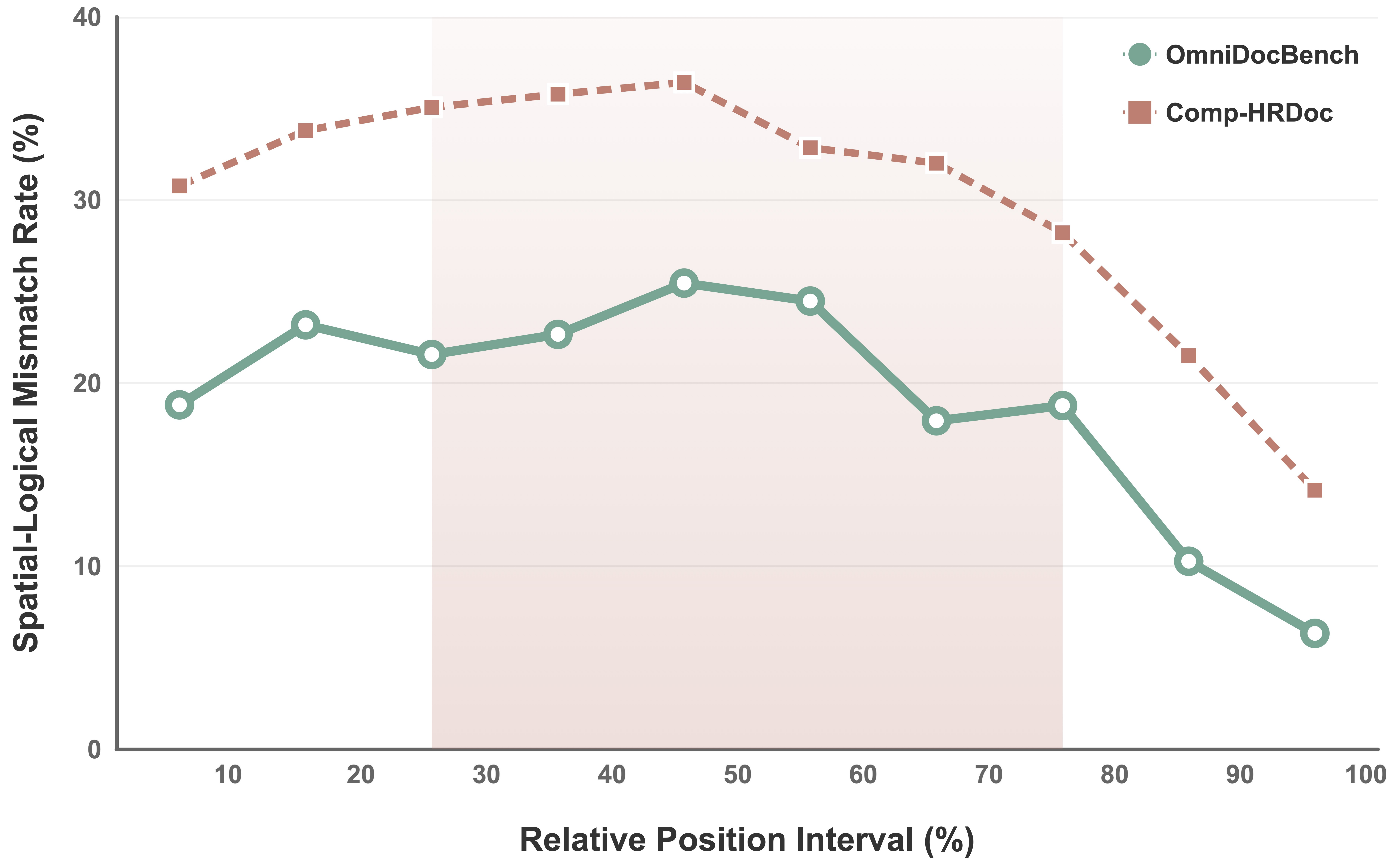}
  \caption{Spatial-Logical Mismatch Analysis. Distribution of spatial-logical mismatches across relative positions on OmniDocBench v1.0 and Comp-HRDoc.}
  \label{fig:mismatch_stat}
\end{figure}

To quantitatively verify the existence of Structural Ambiguity, we introduce a geometric proxy metric: the Spatial-Logical Mismatch.
Specifically, we quantify the density of such mismatches, defined as transitions where the ground-truth next region deviates from the geometrically nearest neighbor.
To ensure reproducibility, we explicitly define the nearest neighbor based on the Euclidean distance between the center points of the respective bounding boxes.
We conduct a geometric analysis on OmniDocBench v1.0 and Comp-HRDoc.
As shown in Figure~\ref{fig:mismatch_stat}, the distribution of these mismatches peaks significantly within the intermediate sections ($20\%$--$80\%$), exhibiting a strong correlation with the error curve.

This correlation exposes a fundamental mismatch between the task's intrinsic complexity and the standard optimization formulation.
Formally, regardless of the architecture, existing methods predominantly optimize the conditional probability of the sequence $Y$ via the standard Cross-Entropy (CE) loss:
\begin{figure*}[t]
  \centering
  \setlength{\abovecaptionskip}{0.12cm}
  \setlength{\belowcaptionskip}{-0.4cm}
  \includegraphics[width=1\textwidth]{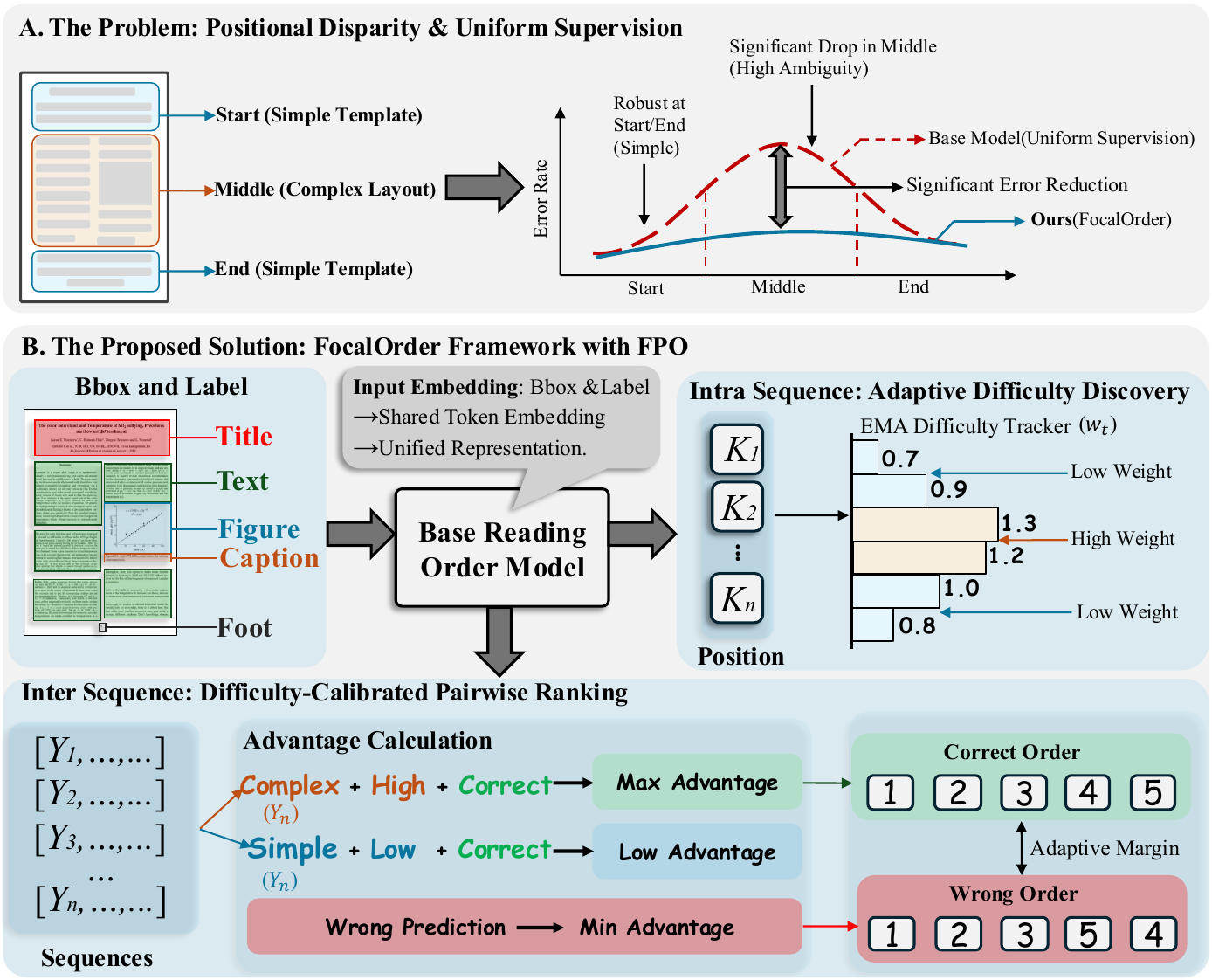}
  \caption{Overview of the FocalOrder framework. The architecture integrates two components: Adaptive Difficulty Discovery, which leverages an EMA-based tracker to dynamically identify and up-weight ($w_t$) structurally ambiguous transitions; and Difficulty-Calibrated Pairwise Ranking, which implements contrastive optimization using a difficulty-aware advantage function and adaptive margins to prioritize complex layout patterns over trivial ones.}
  \label{fig:architecture}
\end{figure*}
\begin{equation}
\label{eq:ce_loss}
\small
\mathcal{L}_{\text{CE}} = - \frac{1}{N} \sum_{t=1}^{N} \log P(y_t \mid y_{<t}, \mathcal{O}, \mathcal{I}).
\end{equation}

The limitation of this formulation lies in its implicit assumption of uniformity.
As seen in Eq.~\ref{eq:ce_loss}, the standard objective applies a static weight ($\frac{1}{N}$) to every transition step $t$. It treats a trivial intra-paragraph connection identically to a complex cross-column jump.

This assumption of uniformity fundamentally misaligns with the inherent structure of documents. As supported by insights from GraphDoc~\cite{chen2025graph}, complex logical relations are significantly sparser than simple spatial neighborhoods. Our analysis further reveals that these critical transitions appear with higher frequency in the intermediate sections.
Consequently, under uniform supervision, the optimization landscape is dominated by the massive volume of trivial patterns found at the start and end. This leads to gradient dilution, where the learning signals from high-ambiguity transitions are overwhelmed by the gradients from easy samples. This causes the model to overfit to simple heuristics and fail at the decision boundaries required to resolve structural ambiguity.

\section{Method}
\label{sec:method}

\subsection{Overview}
\label{sec:Overview}
As illustrated in Figure~\ref{fig:architecture}, the FocalOrder framework is designed to bridge the optimization gap caused by uniform supervision. The workflow begins by encoding layout elements (including Bounding Boxes and Text Labels) into a unified representation via the backbone encoder. To explicitly address structural ambiguity, the optimization process is decomposed into two synergistic pathways that map directly to our mathematical formulation:

\textbf{Intra-Sequence Adaptation (Eq.~\ref{eq:ema_update}--\ref{eq:position_weight}):} The \textit{Adaptive Difficulty Discovery} module functions as a dynamic monitor. It tracks the historical error rates of different layout transitions to compute a position-aware difficulty weight $w_t$. This weight is then applied to the token-level supervision, ensuring that the model focuses more on complex regions (e.g., the document body) rather than trivial start/end tokens.
    
\textbf{Inter-Sequence Alignment (Eq.~\ref{eq:advantage}--\ref{eq:brrl_loss}):} The \textit{Difficulty-Calibrated Pairwise Ranking} module introduces a global contrastive objective. By calculating a difficulty-aware advantage $A_i$, it constructs preference pairs and enforces a ranking loss with adaptive margins $m_{ij}$. This ensures that the model not only predicts local tokens correctly but also maintains global logical coherence.

Finally, these two components are unified in the total objective function (Eq.~\ref{eq:final_loss}), jointly penalizing local sorting errors and global structural inconsistencies.

\subsection{Adaptive Difficulty Discovery}
\label{sec:add}

To mitigate the gradient dilution stemming from the dominance of easy samples, we introduce the Adaptive Difficulty Discovery mechanism. We posit that transition difficulty is inherently dynamic rather than static. To capture this, we partition the sequence into $K$ discrete bins and maintain a global difficulty vector $\mathcal{D} \in \mathbb{R}^K$. This vector tracks the historical loss and is updated via Exponential Moving Average (EMA) to ensure stability:
\begin{equation}
\label{eq:ema_update}
\small
\bar{\mathcal{L}}_k^{(\text{iter})} = \gamma \cdot \bar{\mathcal{L}}_k^{(\text{iter}-1)} + (1-\gamma) \cdot \mathcal{L}_{\text{batch}}^{(k)}.
\end{equation}
Here, $\gamma \in [0, 1)$ serves as a momentum coefficient. 
Crucially, we employ a relatively large $\gamma$ to act as a low-pass filter against batch-wise variance. 
Since document layouts exhibit high diversity, the instantaneous loss within a single batch may fluctuate violently due to data sampling rather than actual learning progress. 
A high momentum ensures that $\bar{\mathcal{L}}_k$ captures the \textit{persistent structural difficulty} (i.e., the stable ``Inverted-U'' disparity profile observed in the dataset) rather than transient noise. 
This allows the difficulty weights $w_t$ to evolve smoothly, providing a stable calibration signal that aligns with the global optimization landscape.
Based on this estimation, the dynamic weight $w_t$ for step $t$ is derived proportional to the relative difficulty of its corresponding bin:
\begin{equation}
\label{eq:position_weight}
\small
w_t = \text{Clip}\left(\frac{\bar{\mathcal{L}}_k}{\mu_{\mathcal{D}}}, w_{\min}, w_{\max}\right).
\end{equation}
Here, $\mu_{\mathcal{D}}$ denotes the mean value of the difficulty vector $\mathcal{D}$, acting as a normalization factor to center the weights. The terms $w_{\min}$ and $w_{\max}$ are clipping thresholds. This formulation effectively constructs a \textit{position-aware focal mechanism}, automatically amplifying gradients from structurally ambiguous regions without requiring manual annotations.

\subsection{Difficulty-Calibrated Pairwise Ranking}
\label{sec:dcpr}

While weighted supervision improves local constraints, it lacks a global perspective. To enforce structural consistency, we introduce the Difficulty-Calibrated Pairwise Ranking (DCPR) objective.

\textbf{Reward Function Definition.}
We evaluate the generated sequence $\hat{Y}$ against the ground truth $Y^*$ using a normalized metric based on the inverted Levenshtein Edit Distance:
\begin{equation}
\small
R(\hat{Y}, Y^*) = 1 - \frac{\text{Lev}(\hat{Y}, Y^*)}{\max(|\hat{Y}|, |Y^*|)}.
\end{equation}

\textbf{Difficulty-Calibrated Advantage.}
We define the advantage $A_i$ for the $i$-th sample by integrating a difficulty bonus into the reward. This allows the model to differentiate between simple and complex successes:
\begin{equation}
\label{eq:advantage}
\small
A_i = R(\hat{Y}_i, Y_i^*) + \beta \cdot \tilde{\mathcal{L}}_{\text{CE}}^{(i)}.
\end{equation}
In this formulation, $\tilde{\mathcal{L}}_{\text{CE}}^{(i)}$ serves as a normalized proxy for the inherent instance difficulty. Consequently, achieving high rewards on difficult samples yields the maximum advantage, thereby prioritizing the optimization of complex structural patterns.

\textbf{Batch-wise Relative Ranking Loss.}
Adopting a contrastive perspective, we construct training pairs $\mathcal{P}$ from the top and bottom $\rho\%$ of samples sorted by advantage. We minimize the ranking loss to maximize the likelihood gap:
\begin{equation}
\label{eq:brrl_loss}
\small
\mathcal{L}_{\text{Rank}} = \frac{1}{|\mathcal{P}|} \sum_{(i,j) \in \mathcal{P}} \big[ S(\hat{Y}_j) - S(\hat{Y}_i) + m_{ij} \big]_+,
\end{equation}
where $S(\cdot)$ represents the sequence log-probability score, and $[\cdot]_+ = \max(0, \cdot)$ denotes the hinge function.
Crucially, we employ a \textit{topology-aware adaptive margin} $m_{ij}$ derived from the difficulty weights in Section~\ref{sec:add}:
\begin{equation}
\small
m_{ij} = \alpha \cdot \max(\bar{w}^{(i)}, \bar{w}^{(j)}).
\end{equation}
Here, $\alpha$ is a base margin scaling factor, and $\bar{w}^{(i)}$ represents the \textit{structural complexity score} of sequence $i$, computed as the mean of its token-level difficulty weights. This novel formulation ensures that pairs involving complex layouts necessitate a larger probability margin, effectively focusing the alignment process on hard samples.

\subsection{Total Objective Function}
\label{sec:total_loss}

The final training objective synergistically combines the difficulty-weighted supervision and the ranking constraint:
\begin{equation}
\label{eq:final_loss}
\small
\mathcal{L}_{\text{total}} = \sum_{t=1}^{N} w_t \cdot \mathcal{L}_{\text{CE}}^{(t)} + \lambda_{\text{Rank}} \cdot \mathcal{L}_{\text{Rank}},
\end{equation}
where $N$ denotes the total number of tokens in the batch, $\mathcal{L}_{\text{CE}}^{(t)}$ is the standard cross-entropy loss at step $t$, and $\lambda_{\text{Rank}}$ is a hyperparameter balancing the ranking constraint. This hybrid objective allows FocalOrder to leverage the stability of supervised learning while capturing global dependencies via preference ranking.

\section{Experiments}
\label{sec:experiments}

\subsection{Datasets and Evaluation Metrics}
To rigorously evaluate our model's capability in handling complex layouts, we conduct experiments on two challenging benchmarks. 
We first utilize \textbf{OmniDocBench}~\cite{ouyang2024omnidocbench}, covering both the foundational v1.0 (981 pages) and the extended v1.5 (1,355 pages).
These datasets are characterized by extreme element density.
 Notably, v1.5 triples the volume of inline formulas, posing significant challenges for local sorting.
Following standard protocols, we report the \textbf{Edit Distance} to measure the deviation between the predicted sequence and the ground truth. Additionally, we evaluate on \textbf{Comp-HRDoc}~\cite{wang2024detect}, a large-scale dataset consisting of 1,500 documents and nearly 1 million annotated elements.
For this benchmark, we employ the \textbf{Reading Edit Distance Score (REDS)} as the primary metric, reporting performance on both text and graphical regions.

\begin{table}[t]
\small
\setlength{\belowcaptionskip}{-0.4cm}
\resizebox{\linewidth}{!}{%
\begin{tabular}{lcc}
\hline
\textbf{Methods} & \textbf{Text Region REDS} & \textbf{Graphical Region REDS} \\
\hline
DOC-R18 & 93.2 & 86.4 \\
UniHDSA-R18 & 96.4 & 90.6 \\
UniHDSA-R50 & 96.7 & 91.0 \\
\hline
\textbf{FocalOrder (Ours)} & \textbf{97.1} & \textbf{91.1} \\
\hline
\end{tabular}%
}
\caption{Performance comparison on the Comp-HRDoc. Metric: Reading Edit Distance Score (REDS), where higher is better. \textbf{Bold} indicates the best.}
\label{tab:comp_hrdoc_sota}
\end{table}

\subsection{Implementation Details}
For a fair and rigorous comparison, we employ the pre-trained LayoutLMv3-large~\cite{HuangL0LW22} as the unified backbone encoder for all our experiments and ablation studies. During the training phase, the initial learning rate is set to $3 \times 10^{-5}$ with a linear warmup for the first 5\% of steps, followed by cosine decay. The momentum coefficient $\gamma$ for the EMA-based difficulty discovery is set to $0.99$. The model is trained for 50 epochs with a batch size of 24 on NVIDIA RTX 4090 GPUs. All baseline results are reproduced using their official codebases or directly cited from respective papers.

\subsection{Comparison with Existing Approaches}

\textbf{Results on Comp-HRDoc}
The results on the Comp-HRDoc, shown in Table~\ref{tab:comp_hrdoc_sota}, further demonstrate the efficacy of our method in handling topological complexity. FocalOrder achieves the highest scores in both categories: \textbf{97.1\%} REDS on Text Regions and \textbf{91.1\%} REDS on Graphical Regions.
It is worth noting that the improvement is consistent across both text flows and graphical elements. While previous methods like UniHDSA~\cite{wang2025unihdsa} show strong performance, our FocalOrder framework effectively mines hard samples, which are often found in graphical regions or complex tables. This leads to a 0.5\% improvement in graphical region serialization over the previous best method. These results empirically support our claim that the point-wise supervision used in baselines is insufficient for structure-defining transitions.

\begin{table}[t]
\small
\setlength{\belowcaptionskip}{-0.2cm}
\resizebox{\linewidth}{!}{%
\begin{tabular}{llcc}
\hline
\multirow{2}{*}{\textbf{Model Type}} & \multirow{2}{*}{\textbf{Methods}} & \multicolumn{2}{c}{\textbf{Edit ($\downarrow$)}} \\ \cline{3-4}
 &  & \textbf{EN} & \textbf{ZH} \\
\hline
\multirow{8}{*}{\shortstack[l]{Pipeline\\Tools}} & MinerU & 0.079 & 0.292 \\
 & Marker & 0.114 & 0.340 \\
 & Mathpix & 0.108 & 0.304 \\
 & Docling & 0.313 & 0.837 \\
 & Pix2Text & 0.281 & 0.499 \\
 & Unstructured & 0.145 & 0.387 \\
 & OpenParse & 0.595 & 0.641 \\
 & PP-StructureV3 & 0.069 & 0.091 \\
\hline
\multirow{7}{*}{\shortstack[l]{General\\VLMs}} & GPT-4o & 0.128 & 0.251 \\
 & Qwen2-VL-72B & 0.119 & 0.193 \\
 & Qwen2.5-VL-72B & 0.106 & 0.168 \\
 & Gemini-1.5 Pro & 0.049 & 0.121 \\
 & Doubao-1.5-pro & 0.058 & 0.094 \\
 & InternVL2-76B & 0.317 & 0.228 \\
 & InternVL3-78B & 0.095 & 0.161 \\
\hline
\multirow{12}{*}{\shortstack[l]{Expert\\VLMs}} & GOT-OCR & 0.141 & 0.280 \\
 & Nougat & 0.382 & 0.954 \\
 & Mistral OCR & 0.083 & 0.284 \\
 & OLMOCR-sglang & 0.145 & 0.277 \\
 & SmolDocling-256M & 0.227 & 0.522 \\
 & Dolphin & 0.091 & 0.162 \\
 & MinerU 2.0 & 0.069 & 0.118 \\
 & OCRFlux & 0.086 & 0.187 \\
 & MonkeyOCR-pro-3B & 0.100 & 0.185 \\
 & dots.ocr & \underline{0.040} & 0.067 \\
 & PaddleOCR-VL & 0.045 & \underline{0.063} \\
 & MinerU 2.5 & 0.045 & 0.068 \\
\hline
\textbf{Ours} & \textbf{FocalOrder} & \textbf{0.038} & \textbf{0.055} \\
\hline
\end{tabular}%
}
\caption{Performance comparison of reading order detection on the OmniDocBench v1.0. \textbf{Bold} indicates the best, and \underline{underline} indicates the second best.}
\label{tab:omniv1_sota}
\end{table}

\textbf{Results on OmniDocBench}
Table~\ref{tab:omniv1_sota} presents the quantitative comparison on OmniDocBench v1.0. Our FocalOrder achieves state-of-the-art performance, recording an Edit Distance of \textbf{0.038} on English documents and \textbf{0.055} on Chinese documents.
Notably, FocalOrder significantly outperforms General VLMs. For instance, compared to GPT-4o (0.128 on EN) and Gemini-1.5 Pro (0.049 on EN), our specialized structural optimization yields a substantial margin. This highlights that while LLMs possess strong semantic understanding, they still struggle with the precise serialization of spatial coordinates in 2D layouts.
Compared to expert models like MinerU 2.5 (0.045 on EN) and PaddleOCR-VL (0.045 on EN), our method achieves a further reduction in error rates. This improvement is attributed to the Difficulty-Calibrated Pairwise Ranking, which prevents the model from being satisfied with ``mostly correct'' sequences and forces it to resolve subtle ordering ambiguities.

\begin{table}[t]
\small
\setlength{\belowcaptionskip}{-0.3cm}
\resizebox{\linewidth}{!}{%
\begin{tabular}{llcc}
\hline
\textbf{Model Type} & \textbf{Methods} & \textbf{Size} & \textbf{Edit ($\downarrow$)} \\
\hline
\multirow{3}{*}{\shortstack[l]{Pipeline\\Tools}} & PP-StructureV3 & - & 0.073 \\
 & MinerU2-pipeline & - & 0.225 \\
 & Marker-1.8.2 & - & 0.250 \\
\hline
\multirow{5}{*}{\shortstack[l]{General\\VLMs}} & Qwen3-VL-Instruct & 235B & 0.068 \\
 & Gemini-2.5 Pro & - & 0.097 \\
 & Qwen2.5-VL & 72B & 0.102 \\
 & InternVL3.5 & 241B & 0.125 \\
 & GPT-4o & - & 0.148 \\
\hline
\multirow{12}{*}{\shortstack[l]{Expert\\VLMs}} & MonkeyOCR-pro-3B & 3B & 0.128 \\
 & dots.ocr & 3B & 0.053 \\
 & DeepSeek-OCR & 3B & 0.086 \\
 & Nanonets-OCR-s & 3B & 0.108 \\
 & MinerU2-VLM & 0.9B & 0.086 \\
 & olmOCR & 7B & 0.121 \\
 & Dolphin-1.5 & 0.3B & 0.080 \\
 & POINTS-Reader & 3B & 0.145 \\
 & Mistral OCR & - & 0.144 \\
 & OCRFlux & 3B & 0.202 \\
 & PaddleOCR-VL & 0.9B & \textbf{0.043} \\
 & MinerU 2.5 & 1.2B & 0.044 \\
\hline
\textbf{Ours} & \textbf{FocalOrder} & 0.4B & \underline{0.044} \\
\hline
\end{tabular}%
}
\caption{Performance comparison on the OmniDocBench v1.5. \textbf{Bold} indicates the best, and \underline{underline} indicates the second best.}
\label{tab:omniv15_sota}
\end{table}

Table~\ref{tab:omniv15_sota} extends the evaluation to the larger OmniDocBench v1.5. Despite the increased dataset scale and variety, FocalOrder maintains high performance with an Edit Distance of \textbf{0.044}. It remains competitive against large-scale models such as Qwen3-VL-Instruct (0.068) and equals the performance of strong baselines like MinerU 2.5, validating the robustness of our approach across different data distributions.

\subsection{Visualization of Learned Weights}
\label{sec:weight_viz}

To validate the efficacy of Adaptive Difficulty Discovery, we visualize the distribution of learned weights $w_t$ on the OmniDocBench v1.0, as shown in Table~\ref{tab:weight_distribution}.
The resulting weight distribution exhibits an ``Inverted-U'' pattern that mirrors the error curve discussed in Section~\ref{sec:analysis}.
Specifically, the model autonomously attenuates weights in the deterministic start and end regions (dropping to 0.32) while amplifying supervision signals in the ambiguous intermediate sections (peaking at 1.61).
This confirms that FocalOrder successfully prioritizes critical structural boundaries over trivial templates without relying on manual heuristics.

\begin{table}[t]
\centering
\setlength{\belowcaptionskip}{-0.3cm}
\small
\resizebox{0.95\linewidth}{!}{%
\begin{tabular}{ccc}
\hline
\textbf{Relative Position} & \textbf{Weight ($w_t$)} & \textbf{Intensity} \\
\hline
0-10\%   & 0.32 & Low \\
10-20\%  & 0.92 & Medium \\
20-30\%  & 1.11 & High \\
30-40\%  & 1.41 & \textbf{Very High} \\
40-50\%  & \textbf{1.61} & \textbf{Peak} \\
50-60\%  & 1.42 & \textbf{Very High} \\
60-70\%  & \textbf{1.61} & \textbf{Peak} \\
70-80\%  & 0.98 & Medium \\
80-90\%  & 0.73 & Low \\
90-100\% & 0.39 & Low \\
\hline
\end{tabular}%
}
\caption{Visualization of learned difficulty weights.}
\label{tab:weight_distribution}
\vspace{0.2cm}
\end{table}

\subsection{Ablation Study}
To verify the contribution of each component in our FocalOrder framework, we conduct a progressive ablation study on OmniDocBench v1.0. The results are summarized in Table~\ref{tab:ablation_v1}.

\begin{table}[tbp]
\centering
\setlength{\belowcaptionskip}{-0.2cm}
\resizebox{\linewidth}{!}{%
\begin{tabular}{lcccc}
\hline
\multirow{2}{*}{\textbf{Method Configuration}} & \textbf{Size} & \textbf{Latency} & \multicolumn{2}{c}{\textbf{Edit ($\downarrow$)}} \\ \cline{4-5}
 & \textbf{(B)} & \textbf{(ms)} & \textbf{EN} & \textbf{ZH} \\
\hline
Base Model & 0.4 & 12.1 & 0.246 & 0.252 \\
+ Fine-tuning & 0.4 & 12.1 & 0.119 & 0.168 \\
+ Category Token Embedding & 0.4 & 12.3 & 0.078 & 0.096 \\
\hline
\textbf{+ Preference Optimization (Standard)} & 0.4 & 12.3 & 0.040 & 0.068 \\
\textbf{+ Preference (EMA Fine-grained Loss)} & 0.4 & 12.4 & 0.045 & 0.058 \\
\textbf{+ Preference (Group Contrastive + EMA)} & 0.4 & 12.3 & \textbf{0.038} & \textbf{0.055} \\
\hline
\end{tabular}%
}
\caption{Ablation study on OmniDocBench v1.0. We progressively integrate components of our framework into the base model. \textbf{Bold} indicates the best.}
\label{tab:ablation_v1}
\end{table}

\textbf{Effectiveness of Preference Optimization.}
Starting from the naive LayoutReader baseline (Row 1), adding fine-tuning and category embeddings (Row 3) brings the Edit Distance down to 0.078 (EN). Introducing a standard PO objective, which uses a standard reward without difficulty calibration (Row 4), significantly improves performance to 0.040. This confirms that sequence-level preference alignment mitigates exposure bias.

\textbf{Impact of Adaptive Difficulty Discovery.}
Replacing the standard PO loss with our EMA-based Fine-grained Loss (Row 5) slightly degrades performance compared to the best standard setting in English but notably improves stability in Chinese (0.058). This suggests that while re-weighting helps, local point-wise weighting alone is insufficient to fully capture global coherence.

\textbf{Impact of Difficulty-Calibrated Pairwise Ranking.}
The full FocalOrder framework (Row 6), which integrates the Adaptive Difficulty Discovery with the Group Contrastive Pairwise Ranking, achieves the best performance (0.038 EN / 0.055 ZH). This indicates that the synergy between identifying hard samples (via EMA) and forcing the model to rank better relative to those difficulties (via Pairwise Ranking) is crucial. The combination effectively shifts the optimization focus from dominant easy transitions to the critical structural boundaries that define layout logic.

\textbf{Inference Efficiency Analysis.}
As indicated in the ``Size'' and ``Latency'' columns of Table~\ref{tab:ablation_v1}, our FocalOrder introduces negligible computational overhead during inference. Since the Difficulty Discovery and Pairwise Ranking modules operate during training, the model structure at test time remains consistent with the base LayoutLMv3 backbone. The marginal increase in latency (from 12.1 ms to 12.3 ms) is primarily attributed to the introduction of additional category token embeddings. This confirms that FocalOrder achieves structural optimization without sacrificing the efficiency required for industrial applications.

\begin{figure}[t]
  \centering
  \setlength{\abovecaptionskip}{0.12cm}
  \setlength{\belowcaptionskip}{-0.3cm}
  \includegraphics[width=\columnwidth]{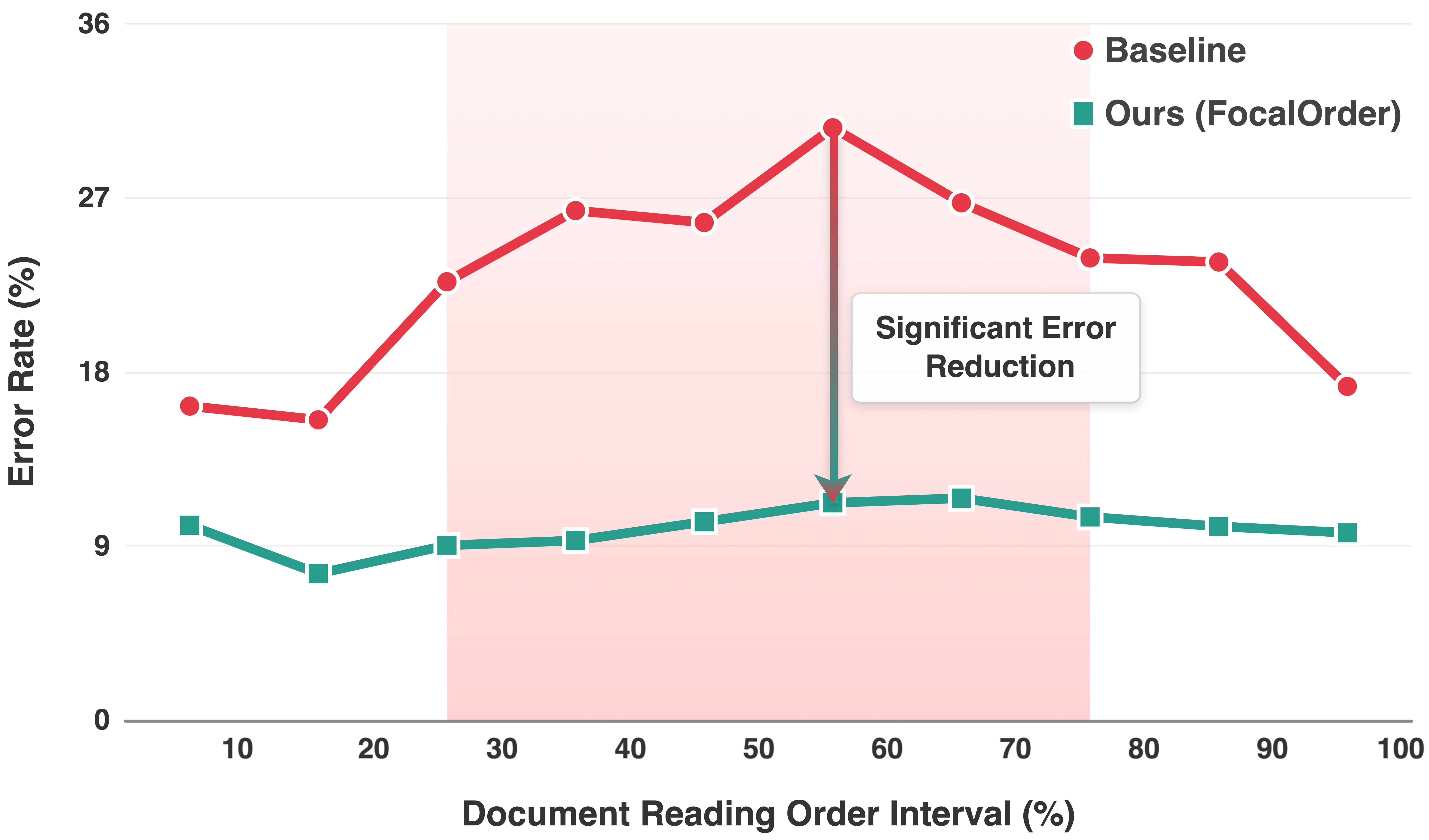}
  \caption{Comparison of error distributions on OmniDocBench v1.0. Unlike the baseline, which suffers from the ``Inverted-U'' degradation, FocalOrder (green line) effectively flattens the curve, maintaining robust performance even in the complex intermediate sections.}
  \label{fig:error_curve}
\end{figure}
\textbf{Mitigating Positional Disparity.}
As visualized in Figure~\ref{fig:error_curve}, the baseline model suffers from a severe ``Inverted-U'' degradation, peaking at \textbf{30.61\%} error in the 50\%--60\% interval. In contrast, FocalOrder effectively flattens this curve by handling structural ambiguity. Specifically, in the intermediate regions (20\%--80\%), our method reduces the average error from 25.99\% to \textbf{10.28\%}, achieving a \textbf{60.4\%} relative improvement. This confirms that our Difficulty-Aware mechanism forces the model to master critical decision boundaries rather than overfitting to trivial templates. Consequently, this yields consistent serialization performance across the entire document, effectively eliminating positional bias.

\textbf{Sensitivity Analysis}
We investigate the impact of the number of difficulty bins $K$ in the Adaptive Difficulty Discovery module. Table~\ref{tab:sensitivity} shows the Edit Distance on OmniDocBench v1.0 with varying $K$.
The model is robust to $K$. $K=1$ degrades to static weighting, yielding suboptimal results. Performance peaks at $K=10$, aligning with the intuition that separating the sequence into deciles effectively captures the rhythm of document layouts, such as differentiating headers, body text, and footers. Excessive granularity ($K=50$) introduces noise, slightly reducing performance.

\begin{table}[t]
\centering
\setlength{\abovecaptionskip}{0.12cm}
\setlength{\belowcaptionskip}{-0.4cm}
\small
\setlength{\tabcolsep}{5.5pt} 
\begin{tabular}{l|ccccc}
\hline
\textbf{Bins} ($K$) & 1 & 5 & 10 & 20 & 50 \\
\hline
\textbf{Edit (EN)} ($\downarrow$) & 0.045 & 0.040 & \textbf{0.038} & 0.039 & 0.041 \\
\textbf{Edit (ZH)} ($\downarrow$) & 0.058 & 0.058 & \textbf{0.055} & 0.055 & 0.056 \\
\hline
\end{tabular}
\caption{Sensitivity analysis on OmniDocBench v1.0. \textbf{Bold} indicates the best.}
\label{tab:sensitivity}
\end{table}

\section{Conclusion}
In this work, we introduce FocalOrder to enhance the reliability of reading order detection in complex document layouts. Rather than relying on standard uniform supervision, which implicitly treats all layout transitions as equally learnable, FocalOrder reframes the problem as a difficulty-aware optimization task, leveraging Adaptive Difficulty Discovery to dynamically prioritize structurally ambiguous regions. Additionally, we propose a Difficulty-Calibrated Pairwise Ranking objective, which adjusts learning margins based on historical error rates to enforce global logical consistency against local noise. Extensive experiments across OmniDocBench v1.0 and Comp-HRDoc demonstrate that FocalOrder effectively flattens the ``Inverted-U'' error curve while establishing new state-of-the-art performance. Notably, our method demonstrates exceptional parameter efficiency for the specific task of layout serialization, achieving superior performance with significantly fewer parameters than massive counterparts. Furthermore, the underlying principle of FocalOrder offers a scalable paradigm for the broader field; future work will explore integrating this difficulty-aware preference mechanism into more general multimodal learning frameworks to further advance visual document understanding.

\section*{Limitations}
This work is presented in light of several limitations regarding the scope and dependencies of our approach. 

Notably, FocalOrder operates as a downstream serialization module contingent upon the granularity of upstream Document Layout Analysis (DLA). Consequently, the model cannot rectify topological errors where layout elements are missed or inaccurately segmented by the preceding detection stage. 

Regarding generalizability, our implementation incorporates semantic category embeddings aligned with the specific ontology of our training benchmarks (English and Chinese). This design choice implies that direct zero-shot application to documents with significantly different semantic schemas or scripts may be constrained, likely necessitating the re-alignment of the embedding space.

We also acknowledge that the definition of a ``correct'' reading order in highly unstructured or artistic layouts retains a degree of subjectivity. Thus, our difficulty-aware formulation may not fully cover all edge cases where the reading path is ambiguous or non-canonical. 

Finally, due to the introduction of the pairwise ranking objective, the training phase incurs a marginal computational overhead compared to standard cross-entropy optimization, though inference latency remains unaffected.

\section*{Ethical Considerations}
We utilize publicly available benchmarks (OmniDocBench and Comp-HRDoc) to conduct the experiments in this study. We adhere to the usage licenses of these datasets and do not anticipate privacy risks, as the data consists of public domain documents.

Since reading order detection is a fundamental capability for automated document understanding, there are dual-use implications. On one hand, precise serialization is pivotal for the reliability of downstream knowledge extraction systems. It ensures that content from complex layouts is fed into RAG pipelines with its original logical coherence preserved, thereby reducing hallucinations caused by disjointed context. On the other hand, improved document parsing capabilities could theoretically be employed by commercial or state actors to facilitate the automated scraping and surveillance of private documents at scale. We do not condone the use of this technology for malicious data mining or privacy infringement. The primary goal of this research is to advance the interpretability and utility of document intelligence systems for public benefit.

\bibliography{custom}

\begin{thebibliography}{21}
\providecommand{\natexlab}[1]{#1}

\bibitem[{Chen et~al.(2025)Chen, Liu, Zheng, Wen, Peng, Zhang, and Stiefelhagen}]{chen2025graph}
Yufan Chen, Ruiping Liu, Junwei Zheng, Di~Wen, Kunyu Peng, Jiaming Zhang, and Rainer Stiefelhagen. 2025.
\newblock \href {https://openreview.net/forum?id=Fu0aggezN9} {Graph-based document structure analysis}.
\newblock In \emph{The Thirteenth International Conference on Learning Representations (ICLR)}.

\bibitem[{Cui et~al.(2025)Cui, Sun, Liang, Gao, and {others}}]{cui2025paddleocr}
Cheng Cui, Ting Sun, Suyin Liang, Tingquan Gao, and {others}. 2025.
\newblock \href {https://arxiv.org/abs/2510.14528} {Paddleocr-vl: Boosting multilingual document parsing via a 0.9b ultra-compact vision-language model}.
\newblock \emph{Preprint}, arXiv:2510.14528.

\bibitem[{Cui et~al.(2021)Cui, Xu, Lv, and Wei}]{cui2021document}
Lei Cui, Yiheng Xu, Tengchao Lv, and Furu Wei. 2021.
\newblock \href {https://arxiv.org/abs/2111.08609} {Document {AI:} benchmarks, models and applications}.
\newblock \emph{Preprint}, arXiv:2111.08609.

\bibitem[{Giovannini and Marinai(2025)}]{giovannini2025survey}
Simone Giovannini and Simone Marinai. 2025.
\newblock \href {https://openaccess.thecvf.com/content/ICCV2025W/VisionDocs/papers/Giovannini_A_Survey_on_Reading_Order_Table_of_Contents_and_Structure_ICCVW_2025_paper.pdf} {A survey on reading order, table of contents, and structure extraction in document analysis}.
\newblock In \emph{Proceedings of the IEEE/CVF International Conference on Computer Vision (ICCV) Workshops}, pages 7585--7594.

\bibitem[{Huang et~al.(2022)Huang, Lv, Cui, Lu, and Wei}]{HuangL0LW22}
Yupan Huang, Tengchao Lv, Lei Cui, Yutong Lu, and Furu Wei. 2022.
\newblock \href {https://dl.acm.org/doi/10.1145/3503161.3548112} {Layoutlmv3: Pre-training for document {AI} with unified text and image masking}.
\newblock In \emph{Proceedings of the 30th ACM International Conference on Multimedia (MM)}, pages 4083--4091.

\bibitem[{Ke et~al.(2025)Ke, Zheng, Li, Xu, Nie, Wang, and He}]{ke2025large}
Wenjun Ke, Yifan Zheng, Yining Li, Hengyuan Xu, Dong Nie, Peng Wang, and Yao He. 2025.
\newblock \href {https://doi.org/10.1145/3768156} {Large language models in document intelligence: A comprehensive survey, recent advances, challenges, and future trends}.
\newblock \emph{ACM Transactions on Information Systems}, 44(1):18:1--18:64.

\bibitem[{Li et~al.(2020)Li, Gao, Bu, Wang, Yu, and Zheng}]{li2020end}
Liangcheng Li, Feiyu Gao, Jiajun Bu, Yongpan Wang, Zhi Yu, and Qi~Zheng. 2020.
\newblock \href {https://www.ecva.net/papers/eccv_2020/papers_ECCV/papers/123700086.pdf} {An end-to-end {OCR} text re-organization sequence learning for rich-text detail image comprehension}.
\newblock In \emph{Proceedings of the European Conference on Computer Vision (ECCV)}, pages 85--100.

\bibitem[{Li et~al.(2025{\natexlab{a}})Li, Yang, Liu, Wang, and Zhang}]{li2025dots}
Yumeng Li, Guang Yang, Hao Liu, Bowen Wang, and Colin Zhang. 2025{\natexlab{a}}.
\newblock \href {https://arxiv.org/abs/2512.02498} {dots.ocr: Multilingual document layout parsing in a single vision-language model}.
\newblock \emph{Preprint}, arXiv:2512.02498.

\bibitem[{Li et~al.(2025{\natexlab{b}})Li, Liu, Liu, Ma, Zhang, Zhang, Guo, Zhang, Wang, and Bai}]{monkeyocr}
Zhang Li, Yuliang Liu, Qiang Liu, Zhiyin Ma, Ziyang Zhang, Shuo Zhang, Zidun Guo, Jiarui Zhang, Xinyu Wang, and Xiang Bai. 2025{\natexlab{b}}.
\newblock \href {https://arxiv.org/abs/2506.05218} {Monkeyocr: Document parsing with a structure-recognition-relation triplet paradigm}.
\newblock \emph{Preprint}, arXiv:2506.05218.

\bibitem[{Mathew et~al.(2021)Mathew, Karatzas, and Jawahar}]{MathewKJ21}
Minesh Mathew, Dimosthenis Karatzas, and C.~V. Jawahar. 2021.
\newblock \href {https://openaccess.thecvf.com/content/WACV2021/papers/Mathew_DocVQA_A_Dataset_for_VQA_on_Document_Images_WACV_2021_paper.pdf} {Docvqa: {A} dataset for {VQA} on document images}.
\newblock In \emph{Proceedings of the IEEE/CVF Winter Conference on Applications of Computer Vision (WACV)}, pages 2200--2209.

\bibitem[{Meunier(2005)}]{meunier2005optimized}
Jean{-}Luc Meunier. 2005.
\newblock \href {https://ieeexplore.ieee.org/document/1575567} {Optimized xy-cut for determining a page reading order}.
\newblock In \emph{Eighth International Conference on Document Analysis and Recognition (ICDAR)}, pages 347--351.

\bibitem[{Niu et~al.(2025)Niu, Liu, Gu, Wang, and {others}}]{mineru2.5}
Junbo Niu, Zheng Liu, Zhuangcheng Gu, Bin Wang, and {others}. 2025.
\newblock \href {https://arxiv.org/abs/2509.22186} {Mineru2.5: {A} decoupled vision-language model for efficient high-resolution document parsing}.
\newblock \emph{Preprint}, arXiv:2509.22186.

\bibitem[{Ouyang et~al.(2025)Ouyang, Qu, Zhou, Zhu, and {others}}]{ouyang2024omnidocbench}
Linke Ouyang, Yuan Qu, Hongbin Zhou, Jiawei Zhu, and {others}. 2025.
\newblock \href {https://openaccess.thecvf.com/content/CVPR2025/papers/Ouyang_OmniDocBench_Benchmarking_Diverse_PDF_Document_Parsing_with_Comprehensive_Annotations_CVPR_2025_paper.pdf} {Omnidocbench: Benchmarking diverse {PDF} document parsing with comprehensive annotations}.
\newblock In \emph{Proceedings of the Computer Vision and Pattern Recognition Conference (CVPR)}, pages 24838--24848.

\bibitem[{Qiao et~al.(2024)Qiao, Li, Cheng, Xu, Niu, and Li}]{qiao2024reading}
Liang Qiao, Can Li, Zhanzhan Cheng, Yunlu Xu, Yi~Niu, and Xi~Li. 2024.
\newblock \href {https://doi.org/10.1016/j.patcog.2024.110314} {Reading order detection in visually-rich documents with multi-modal layout-aware relation prediction}.
\newblock \emph{Pattern Recognition}, 150:110314.

\bibitem[{Wang et~al.(2025{\natexlab{a}})Wang, Wu, Li, Fang, Liang, Huang, Wang, Huang, Chen, Chu, and Qi}]{wang2025infinity}
Baode Wang, Biao Wu, Weizhen Li, Meng Fang, Yanjie Liang, Zuming Huang, Haozhe Wang, Jun Huang, Ling Chen, Wei Chu, and Yuan Qi. 2025{\natexlab{a}}.
\newblock \href {https://arxiv.org/abs/2506.03197} {Infinity parser: Layout aware reinforcement learning for scanned document parsing}.
\newblock \emph{Preprint}, arXiv:2506.03197.

\bibitem[{Wang et~al.(2025{\natexlab{b}})Wang, Hu, and Huo}]{wang2025unihdsa}
Jiawei Wang, Kai Hu, and Qiang Huo. 2025{\natexlab{b}}.
\newblock \href {https://arxiv.org/abs/2503.15893} {Unihdsa: {A} unified relation prediction approach for hierarchical document structure analysis}.
\newblock \emph{Pattern Recognition}, 165:111617.

\bibitem[{Wang et~al.(2024)Wang, Hu, Zhong, Sun, and Huo}]{wang2024detect}
Jiawei Wang, Kai Hu, Zhuoyao Zhong, Lei Sun, and Qiang Huo. 2024.
\newblock \href {https://doi.org/10.1016/j.patcog.2024.110836} {Detect-order-construct: {A} tree construction based approach for hierarchical document structure analysis}.
\newblock \emph{Pattern Recognition}, 156:110836.

\bibitem[{Wang et~al.(2021)Wang, Xu, Cui, Shang, and Wei}]{wang2021layoutreader}
Zilong Wang, Yiheng Xu, Lei Cui, Jingbo Shang, and Furu Wei. 2021.
\newblock \href {https://aclanthology.org/2021.emnlp-main.389/} {Layoutreader: Pre-training of text and layout for reading order detection}.
\newblock In \emph{Proceedings of the 2021 Conference on Empirical Methods in Natural Language Processing (EMNLP)}, pages 4735--4744.

\bibitem[{Wei et~al.(2025)Wei, Sun, and Li}]{deepseekocr}
Haoran Wei, Yaofeng Sun, and Yukun Li. 2025.
\newblock \href {https://arxiv.org/abs/2510.18234} {Deepseek-ocr: Contexts optical compression}.
\newblock \emph{Preprint}, arXiv:2510.18234.

\bibitem[{Wu et~al.(2008)Wu, Chou, and Chang}]{wu2008machine}
Chung-Chih Wu, Chien-Hsing Chou, and Fu~Chang. 2008.
\newblock \href {https://www.sciencedirect.com/science/article/pii/S0031320308000976} {A machine-learning approach for analyzing document layout structures with two reading orders}.
\newblock \emph{Pattern Recognition}, 41(10):3200--3213.

\bibitem[{Zhang et~al.(2025)Zhang, Zhang, Wang, Ouyang, Wen, Li, Chow, He, and Zhang}]{zhang2025ocr}
Junyuan Zhang, Qintong Zhang, Bin Wang, Linke Ouyang, Zichen Wen, Ying Li, Ka-Ho Chow, Conghui He, and Wentao Zhang. 2025.
\newblock \href {https://openaccess.thecvf.com/content/ICCV2025/papers/Zhang_OCR_Hinders_RAG_Evaluating_the_Cascading_Impact_of_OCR_on_ICCV_2025_paper.pdf} {Ocr hinders rag: Evaluating the cascading impact of ocr on retrieval-augmented generation}.
\newblock In \emph{Proceedings of the IEEE/CVF International Conference on Computer Vision (ICCV)}, pages 17443--17453.

\end{thebibliography}

\appendix

\section{Mathematical Formalization and Definitions}
\label{sec:math_details}

In this section, we provide precise definitions and formalizations to ensure reproducibility and clarify the metric calculation protocols used in our analysis.

\subsection{Definition of Batch-wise Bin Loss (Eq. 2)}
In Eq. (2), $\mathcal{L}_{\text{batch}}^{(k)}$ represents the average cross-entropy loss for all tokens falling into the $k$-th position bin within the current batch. Let $\mathcal{B}$ denote the current batch. For a sequence of length $T$, the relative position of the $t$-th token is $p_t = t/T$. The index of the bin is determined by $k = \lfloor p_t \cdot K \rfloor$.
The term is calculated as:
\begin{equation}
    \mathcal{L}_{\text{batch}}^{(k)} = \frac{\sum_{(x, y) \in \mathcal{B}} \sum_{t=1}^{T} \mathbb{I}(k_t = k) \cdot \ell_{\text{CE}}(y_t, y_{<t})}{\sum_{(x, y) \in \mathcal{B}} \sum_{t=1}^{T} \mathbb{I}(k_t = k) + \epsilon},
\end{equation}
where $\mathbb{I}(\cdot)$ is the indicator function, $\ell_{\text{CE}}$ is the token-level cross-entropy loss, and $\epsilon$ is a small constant for numerical stability.

\subsection{Clipping Mechanism (Eq. 3)}
To prevent gradient explosion, weights are clipped dynamically:
\begin{equation}
    w_t = \text{Clip}\left(\frac{\bar{\mathcal{L}}_k}{\mu_{\mathcal{D}}}, 1 - \delta, 1 + \delta \right),
\end{equation}
where $\mu_{\mathcal{D}} = \frac{1}{K} \sum_{k=1}^K \bar{\mathcal{L}}_k$. We set $\delta=0.8$, yielding an effective range of $[0.2, 1.8]$.

\subsection{Difficulty-Calibrated Advantage (Eq. 5)}
The advantage function balances sequence quality and instance difficulty:
\begin{equation}
    A_i = R(\hat{Y}_i, Y_i^*) + \beta \cdot \tilde{\mathcal{L}}_{\text{CE}}^{(i)},
\end{equation}
where $R(\cdot)$ is the edit-distance-based reward. $\tilde{\mathcal{L}}_{\text{CE}}^{(i)}$ is the length-normalized sequence loss, further normalized by the global running average loss to ensure scale consistency. We set $\beta=0.05$.
We analyze the potential interaction between reward and loss: the reward term $R \in [0, 1]$ typically dominates the advantage score. The term $\beta \cdot \tilde{\mathcal{L}}_{\text{CE}}^{(i)}$ acts as a tie-breaker to boost hard samples. Purely wrong predictions (low $R$), even with high loss, will still be ranked lower than correct predictions, ensuring optimization stability.

\subsection{Ranking Score (Eq. 6)}
The ranking score $S(\hat{Y})$ is the length-normalized log-probability:
\begin{equation}
    S(\hat{Y}) = \frac{1}{|\hat{Y}|} \sum_{t=1}^{|\hat{Y}|} \log P(y_t | y_{<t}),
\end{equation}
Normalization prevents bias towards shorter sequences, as unnormalized log-probabilities strictly decrease with sequence length.

\subsection{Definition of Position-wise Error Rate (For Fig. 2)}
To rigorously quantify the ``Positional Disparity,'' we define the error rate based on the optimal alignment between the predicted sequence $\hat{Y}$ and the ground truth $Y^*$.
1. We compute the Levenshtein distance between $\hat{Y}$ and $Y^*$.
2. During the backtrace of the dynamic programming matrix, we identify alignment operations (Match, Substitution, Insertion, Deletion).
3. For each position index $t$ in the ground truth $Y^*$, if the operation is a ``Match'', the error is 0; for ``Substitution'' or ``Deletion'', the error is 1. (Insertions are attributed to the preceding ground truth index).
4. These binary error flags are then aggregated into $K=10$ bins based on their relative position $t/|Y^*|$.
This method ensures that the error rate reflects the model's inability to recall the correct element at the specific relative topological position.

\section{Implementation Details}

\subsection{FocalOrder Algorithm Pseudocode}
\label{sec:algo_code}
Algorithm \ref{alg:focalorder} summarizes the training flow, elucidating the interaction between EMA updates, weight calculation, and the ranking objective.

\begin{algorithm}[h]
\caption{FocalOrder Training Step}
\label{alg:focalorder}
\begin{algorithmic}[1]
\Require Batch $\mathcal{B}$, EMA Difficulty Vector $\mathcal{D}$, Momentum $\gamma$
\State \textbf{Forward Pass:}
\State Compute token logits and $\ell_{\text{CE}}$ for all samples in $\mathcal{B}$.
\State \textbf{Adaptive Difficulty Discovery:}
\For{$k=1$ \textbf{to} $K$}
    \State Calculate batch-wise bin loss $\mathcal{L}_{\text{batch}}^{(k)}$ (Eq. A.1).
    \State Update global difficulty: $\mathcal{D}_k \leftarrow \gamma \mathcal{D}_k + (1-\gamma) \mathcal{L}_{\text{batch}}^{(k)}$.
\EndFor
\State Compute weights $w_t$ for each token based on $\mathcal{D}$ (Eq. A.2).
\State $\mathcal{L}_{\text{Weighted\_CE}} = \sum w_t \cdot \ell_{\text{CE}}$.
\State \textbf{Difficulty-Calibrated Pairwise Ranking:}
\State Calculate Advantage $A_i = R_i + \beta \tilde{\mathcal{L}}^{(i)}$.
\State Sort $\mathcal{B}$ by $A_i$.
\State Select $\mathcal{P}_{\text{pos}}$ (top $\rho\%$) and $\mathcal{P}_{\text{neg}}$ (bottom $\rho\%$).
\State Sample pairs $(i, j)$ from $\mathcal{P}_{\text{pos}} \times \mathcal{P}_{\text{neg}}$.
\State Calculate margin $m_{ij} = \alpha \cdot \max(\bar{w}^{(i)}, \bar{w}^{(j)})$.
\State $\mathcal{L}_{\text{Rank}} = \frac{1}{|\text{pairs}|} \sum \max(0, S_j - S_i + m_{ij})$.
\State \textbf{Update:}
\State $\mathcal{L}_{\text{total}} = \mathcal{L}_{\text{Weighted\_CE}} + \lambda_{\text{Rank}} \mathcal{L}_{\text{Rank}}$.
\State Backward pass and optimizer step.
\end{algorithmic}
\end{algorithm}

\subsection{Details on Inputs and Category Embeddings}
\label{sec:category_fairness}
To ensure a fair comparison, all experiments (including baselines and FocalOrder) utilize the same input features.
\textbf{Category Inputs:} The ``Category Token Embeddings'' refer to the semantic class of the layout element (e.g., ``Text'', ``Title'', ``Figure'', ``Table''). These category labels are provided as part of the input sequence.
\textbf{Fairness:} We do \textbf{not} use Ground Truth categories during inference if they are not available to the baselines. The category inputs are assumed to be obtained from the upstream layout analysis model (e.g., a detection model). Since the same input setting is applied to all compare methods (Baseline, Fine-tuning, FocalOrder), the performance gains reported in Table 5 are strictly due to the proposed optimization strategy.

\subsection{Data Availability and Reproducibility.} 
Due to the upload size limitations of the submission system, we have included only a representative subset of the training data in the supplementary materials.

\section{Extended Analysis and Robustness}

\subsection{Comparison with Simple Baselines}
\label{sec:comparison_baselines}
To investigate whether the performance improvement stems from the \textit{dynamic} EMA mechanism or simply from \textit{any} non-uniform weighting, we compared FocalOrder against a ``Static Inverted-U'' baseline, where weights are manually fixed to follow a Gaussian-like curve (low at ends, high in middle).

\begin{table}[h]
\centering
\small
\begin{tabular}{lc}
\hline
\textbf{Method} & \textbf{Edit Distance ($\downarrow$)} \\
\hline
Uniform Supervision (Baseline) & 0.045 \\
Static Inverted-U Weighting & 0.042 \\
Token-level EMA Weighting & 0.043 \\
\textbf{FocalOrder (Bin-level EMA)} & \textbf{0.038} \\
\hline
\end{tabular}
\caption{Comparison with alternative weighting strategies.}
\label{tab:reweighting_baseline}
\end{table}

As shown in Table~\ref{tab:reweighting_baseline}, while Static Weighting offers a slight improvement over the uniform baseline, it underperforms compared to FocalOrder.
This limitation arises because static heuristics (e.g., a fixed Gaussian curve) impose a rigid prior that may not perfectly align with the \textbf{actual error distribution} of the data.
In contrast, our EMA-based approach is \textbf{data-driven}, allowing the optimization landscape to adaptively fit the intrinsic difficulty profile of the dataset.

Furthermore, ``Token-level EMA'', where difficulty is tracked per-token without spatial binning, yields suboptimal results ($0.043$).
We attribute this to the fact that point-wise error signals are highly susceptible to \textbf{label noise and the inherent subjectivity of reading order} (e.g., ambiguous floating figures).
In this context, Binning ($K=10$) acts as a critical \textbf{regularizer}.
By aggregating statistics over spatial regions, it filters out instance-specific outliers and forces the model to focus on robust \textbf{regional structural ambiguity} rather than overfitting to noisy annotations.

\subsection{Hyperparameter Sensitivity Analysis}
\label{sec:sensitivity}
We analyze the sensitivity of FocalOrder to key hyperparameters on OmniDocBench v1.0 (EN).

\textbf{Sensitivity to $\beta$ (Advantage Weight):}
The parameter $\beta$ controls the contribution of difficulty to the advantage score.

\begin{table}[h]
\centering
\small
\setlength{\abovecaptionskip}{0.1cm}
\setlength{\belowcaptionskip}{-0.1cm}
\begin{tabular}{l|ccccc}
\hline
$\beta$ & 0.0 & 0.01 & \textbf{0.05} & 0.1 & 0.2 \\
\hline
Edit ($\downarrow$) & 0.040 & 0.039 & \textbf{0.038} & 0.039 & 0.042 \\
\hline
\end{tabular}
\caption{Sensitivity analysis of the advantage weight $\beta$.}
\label{tab:sensitivity_beta}
\end{table}

Setting $\beta=0$ reduces the method to standard reward-based ranking. A moderate $\beta=0.05$ yields the best results. Large $\beta$ ($0.2$) leads to performance degradation. This indicates that while incorporating difficulty improves learning, the reward signal (sequence correctness) must remain the dominant factor in the advantage function. However, the method remains stable within the range $[0.01, 0.1]$.

\textbf{Sensitivity to $\rho$ (Pair Selection Ratio):}

\begin{table}[h]
\centering
\small
\setlength{\abovecaptionskip}{0.1cm}
\setlength{\belowcaptionskip}{-0.1cm}
\begin{tabular}{l|ccc}
\hline
$\rho$ & 10\% & \textbf{20\%} & 30\% \\
\hline
Edit ($\downarrow$) & 0.039 & \textbf{0.038} & 0.040 \\
\hline
\end{tabular}
\caption{Sensitivity analysis of the pair selection ratio $\rho$.}
\label{tab:sensitivity_rho}
\end{table}

A ratio of $\rho=20\%$ provides a balanced set of hard positives and negatives.

\section{Qualitative Visualization}
\label{sec:qualitative_viz}

To intuitively demonstrate the efficacy of FocalOrder, we provide a detailed visual comparison on the OmniDocBench dataset.
The visualization includes the original image, as well as predictions from our method, PaddleOCR-VL, and MinerU 2.5.

\begin{figure*}[t]
  \centering
  \setlength{\abovecaptionskip}{0.12cm}
  \setlength{\belowcaptionskip}{-0.2cm}
  \includegraphics[width=\textwidth]{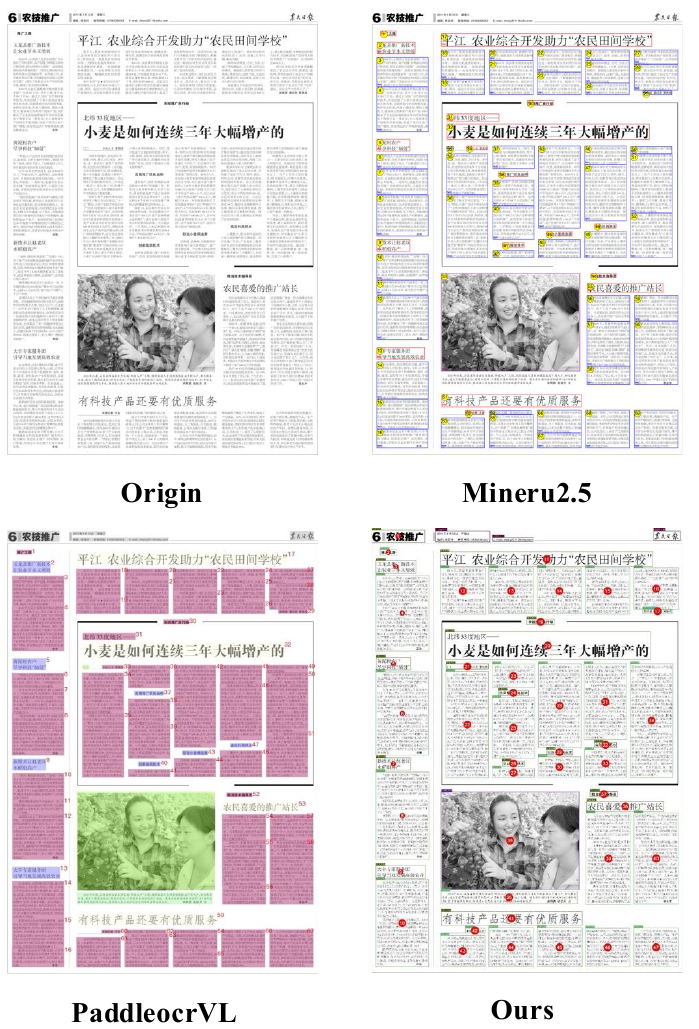}
  \caption{Qualitative comparison of reading order detection on a newspaper layout.}
  \label{fig:appendix_viz1}
\end{figure*}

\begin{figure*}[t]
  \centering
  \setlength{\abovecaptionskip}{0.12cm}
  \setlength{\belowcaptionskip}{-0.2cm}
  \includegraphics[width=\textwidth]{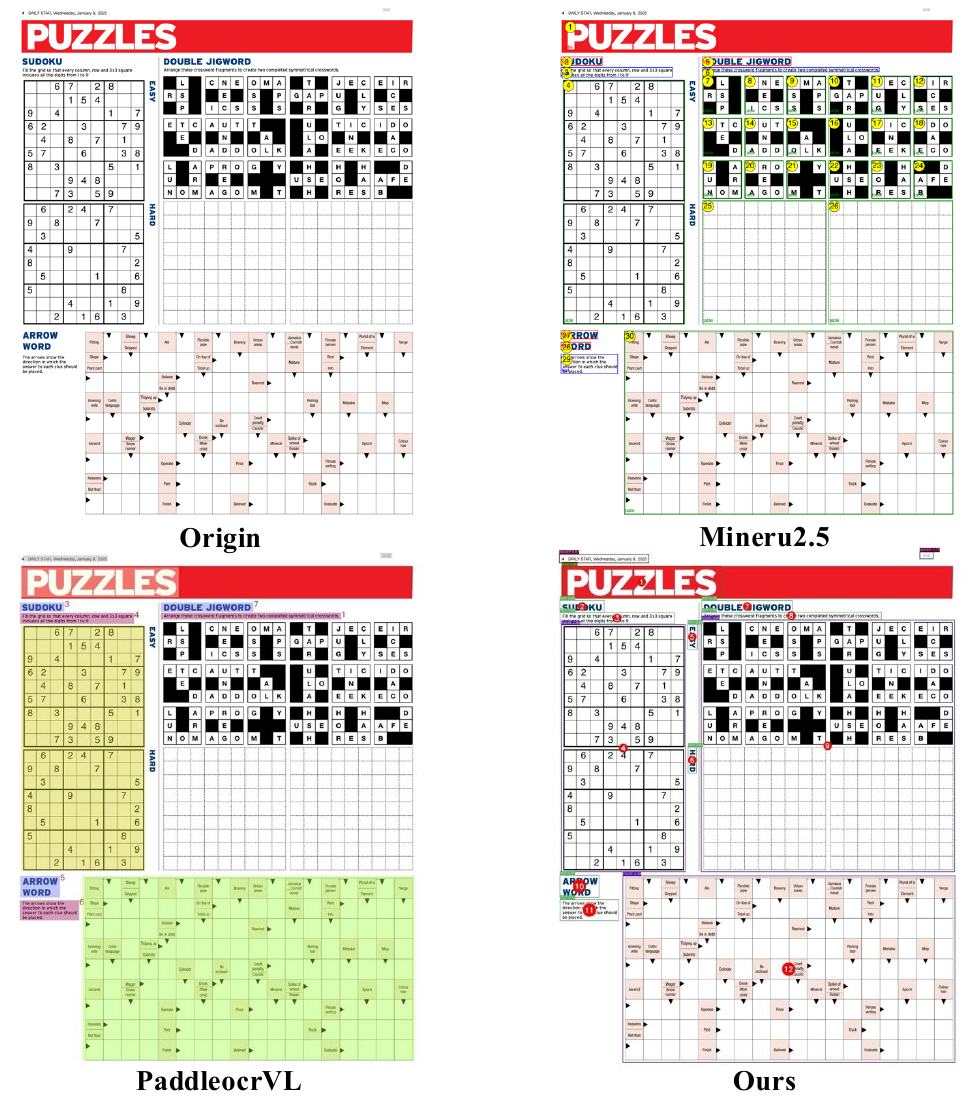}
  \caption{Qualitative comparison of reading order detection on an irregular magazine layout.}
  \label{fig:appendix_viz2}
\end{figure*}

\begin{figure*}[t]
  \centering
  \setlength{\abovecaptionskip}{0.12cm}
  \setlength{\belowcaptionskip}{-0.2cm}
  \includegraphics[width=\textwidth]{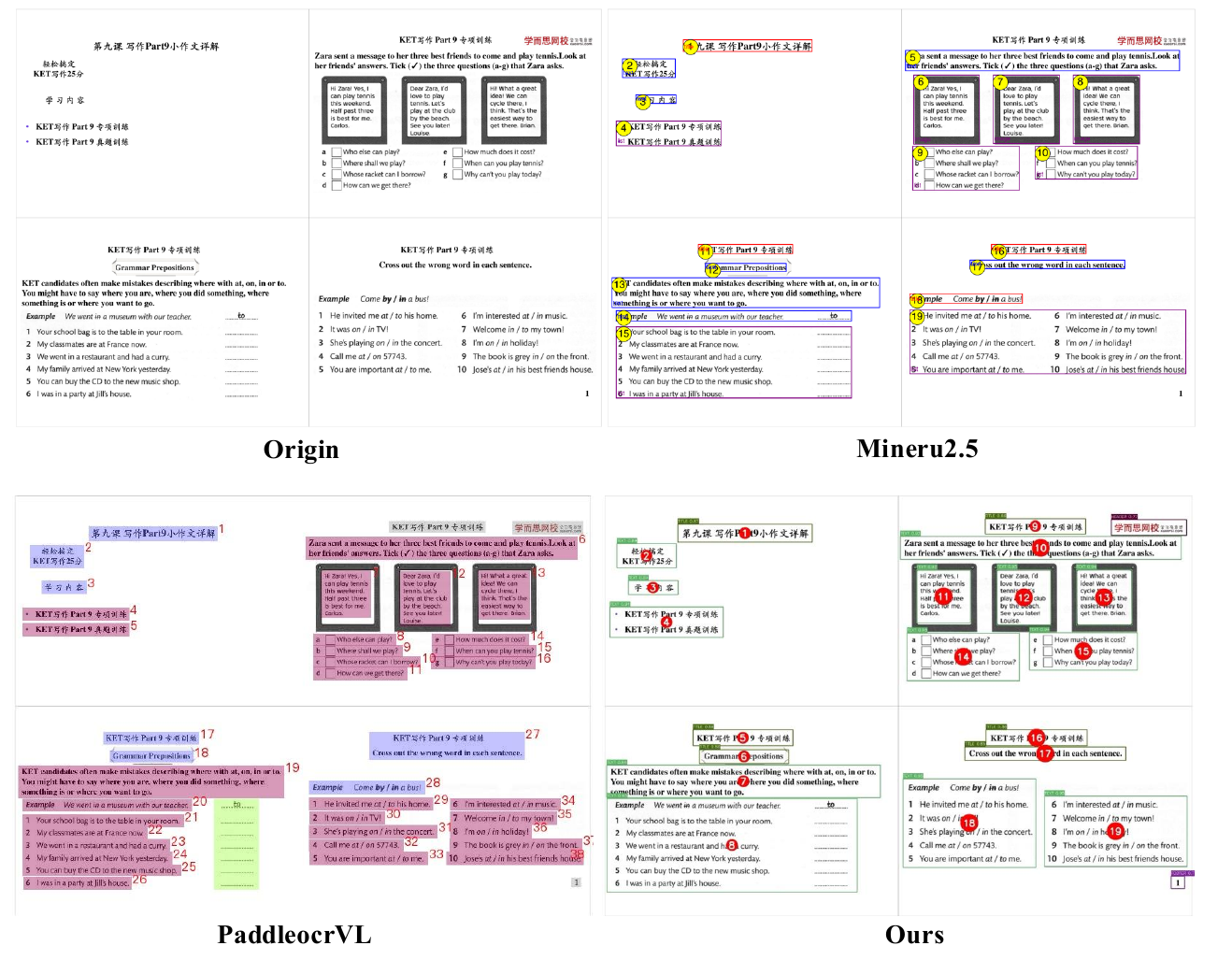}
  \caption{Qualitative comparison of reading order detection on a courseware slide.}
  \label{fig:appendix_viz3}
\end{figure*}

\begin{figure*}[t]
  \centering
  \setlength{\abovecaptionskip}{0.12cm}
  \setlength{\belowcaptionskip}{-0.2cm}
  \includegraphics[width=0.95\textwidth]{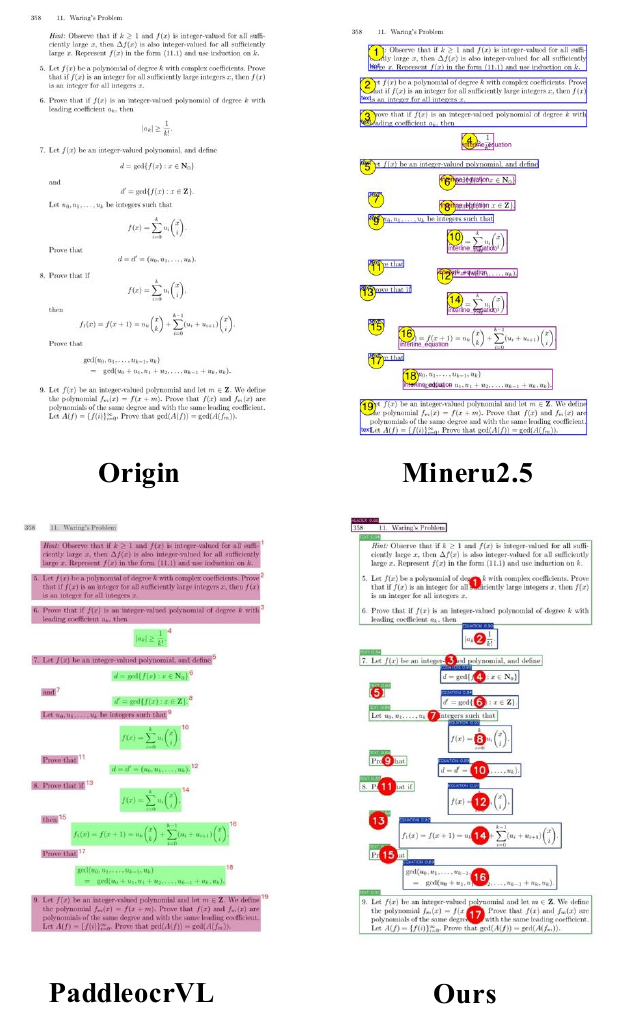}
  \caption{Qualitative comparison of reading order detection on a scientific document with equations.}
  \label{fig:appendix_viz4}
\end{figure*}

\begin{figure*}[t]
  \centering
  \setlength{\abovecaptionskip}{0.12cm}
  \setlength{\belowcaptionskip}{-0.2cm}
  \includegraphics[width=\textwidth]{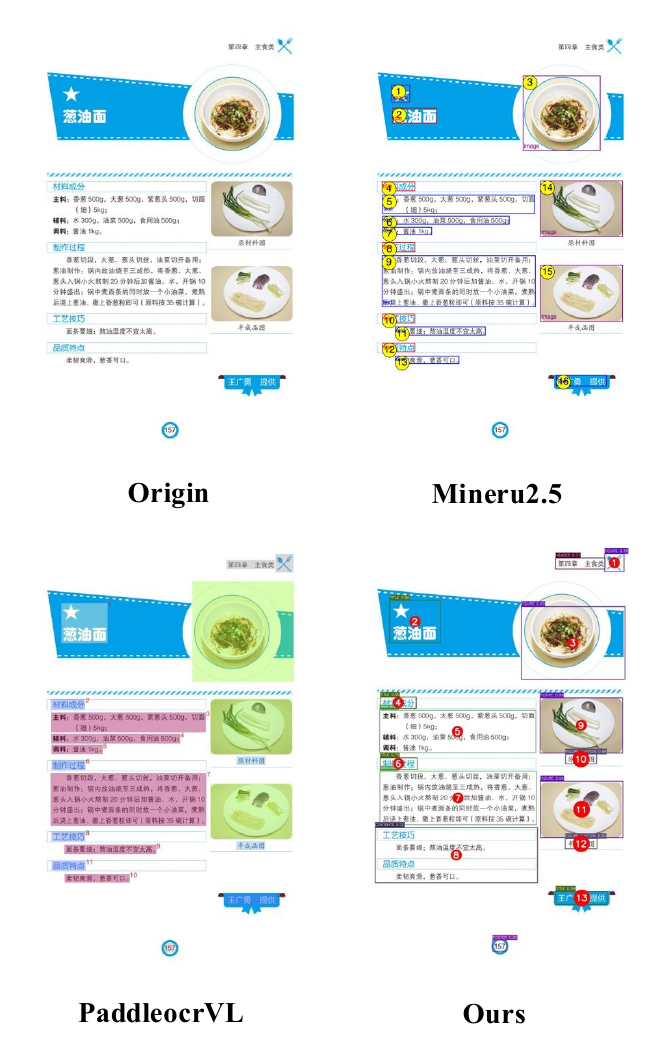}
  \caption{Qualitative comparison of reading order detection on a textbook page with text wrapping.}
  \label{fig:appendix_viz5}
\end{figure*}

As illustrated in Figures~\ref{fig:appendix_viz1}--\ref{fig:appendix_viz5}, facing reading order prediction under complex layout samples, our method significantly outperforms PaddleOCR-VL, which utilizes pointer networks.
Furthermore, FocalOrder demonstrates comparable performance to MinerU 2.5, which employs a multi-stage VLM pipeline, with both methods showing competitive results on challenging cases.
These observations empirically validate the feasibility and robustness of our proposed approach.

\section{AI Usage Declaration}
\label{sec:ai_usage}

We acknowledge the use of AI assistants for grammatical polishing to ensure linguistic clarity. We strictly adhered to the ACL 2026 policies regarding AI assistance:
\begin{itemize}
    \item The AI tool was used solely for improving the readability, flow, and grammatical correctness of the text.
    \item No scientific claims, experimental results, or core ideas were generated by the AI.
    \item All outputs from the model were manually verified and revised by the authors to ensure accuracy.
\end{itemize}
The authors accept full responsibility for the content of this paper.

\end{document}